\pdfoutput=1

\documentclass[11pt]{article}

\usepackage[]{emnlp2021}

\usepackage{times}
\usepackage{latexsym}

\usepackage{microtype}
\usepackage{booktabs}
\usepackage{multirow}
\usepackage{graphicx}
\usepackage{subcaption}
\usepackage[export]{adjustbox}
\usepackage{hyperref}
\usepackage{tabularx}
\usepackage{paralist}
\usepackage{listings}
\usepackage{float}
\usepackage{placeins}
\usepackage{tikz}
\def\checkmark{\tikz\fill[scale=0.4](0,.35) -- (.25,0) -- (1,.7) -- (.25,.15) -- cycle;} 

\usepackage[T1]{fontenc}

\usepackage[utf8]{inputenc}

\usepackage{microtype}

\title{Vyākarana: A Colorless Green Benchmark for Syntactic Evaluation in Indic Languages}

\author{Rajaswa Patil\\
  Cognitive Neuroscience Lab\\
  BITS Pilani, Goa Campus\\
  \texttt{\scriptsize f20170334@goa.bits-pilani.ac.in}\\ \And
  
  Jasleen Dhillon\\
  Dept. of CS \& IS\\
  BITS Pilani, Goa Campus\\
  \texttt{\scriptsize f20170077@goa.bits-pilani.ac.in} \\ \And
  
  Siddhant Mahurkar\\
  Data Science Institute \\
  Columbia University\\
  \texttt{\scriptsize sm5129@columbia.edu} \\ \AND
  
  Saumitra Kulkarni\\
  Dept. of Computer Engineering\\
  PICT, Maharashtra, India\\
  \texttt{\scriptsize kulkarnisaumitra98@gmail.com}\\ \And
  
  Manav Malhotra\\
  Dept. of Mathematics\\
  BITS Pilani, Goa Campus\\
  \texttt{\scriptsize f20171016@goa.bits-pilani.ac.in} \\ \And
  
  Veeky Baths\\
  Cognitive Neuroscience Lab\\
  BITS Pilani, Goa Campus\\
  \texttt{\scriptsize veeky@goa.bits-pilani.ac.in} \\ 
}

\begin{document}
\maketitle
\begin{abstract}
While there has been significant progress towards developing NLU resources for Indic languages, syntactic evaluation has been relatively less explored. Unlike English, Indic languages have rich morphosyntax, grammatical genders, free linear word-order, and highly inflectional morphology. In this paper, we introduce \emph{Vyākarana}: a benchmark of Colorless Green sentences in Indic languages for syntactic evaluation of multilingual language models. The benchmark comprises four syntax-related tasks: PoS Tagging, Syntax Tree-depth Prediction, Grammatical Case Marking, and Subject-Verb Agreement. We use the datasets from the evaluation tasks to probe five multilingual language models of varying architectures for syntax in Indic languages. Due to its prevalence, we also include a code-switching setting in our experiments. Our results show that the token-level and sentence-level representations from the Indic language models (IndicBERT and MuRIL) do not capture the syntax in Indic languages as efficiently as the other highly multilingual language models. Further, our layer-wise probing experiments reveal that while mBERT, DistilmBERT, and XLM-R localize the syntax in middle layers, the Indic language models do not show such syntactic localization.
\end{abstract}

\section{Introduction}
\label{section-introduction}
The Indian subcontinent is home to more than $450$ languages spanning seven language families. Multilingualism and code-switching are common phenomenon across these languages.  Traditionally, the region has been at the center of many linguistic studies due to its rich linguistic diversity \citep{10.2307/410649}. The Indo-Aryan and the Dravidian language families are the most prominent ones in the subcontinent, with more than a billion speakers combined. Recent work in computational linguistics has focused on clubbing together the major languages from these two language families into a single group called as \emph{Indic Languages}.\footnote{From a linguistic perspective, the term \emph{Indic} is used for the Indo-Aryan language family. However, in this work, we use the term to represent all the languages used in the subcontinent (following the recent work in ``Indic NLP").} Even with such a huge potential user base, the progress in building language technologies for the Indic languages has been limited. Most of the Indic languages fall under the category of low-resourced and mid-resourced languages. Some recent open-sourced efforts have tried to address this by developing various toolkits \citep{arora-2020-inltk}, frameworks \citep{kunchukuttan2020indicnlp}, language models, benchmarks, and datasets \citep{kakwani-etal-2020-indicnlpsuite}. 

To perform well on natural language understanding tasks, a language model should have a good understanding of the various general aspects of the structure of language, like its grammar and syntax. The structure of Indic languages is quite rich in its morphosyntax, which is partially shared across the languages. Thus, performing language modeling and downstream tasks on a multilingual dataset of Indic languages helps capture typologically generalized stimuli across the languages and implicitly addresses the issue of data-scarcity. While there have been some efforts to analyze the role of multilingual training for Indic languages, it has been limited to downstream NLU tasks and large multilingual models which are not exclusively trained for Indic languages \citep{jain2020indictransformers}. In this work, we introduce a syntactic evaluation benchmark of \emph{Colorless Green}\footnote{\emph{Colorless green} sentences are grammatically correct, but semantically nonsensical.} sentences in Indic languages: \emph{Vyākarana}\footnote{Vyākarana is the Sanskrit term for grammar}, which provides a challenging environment to evaluate multilingual language models for their syntactic abilities specifically. We use the dataset to conduct layer-wise probing of five multilingual language models to inspect their understanding of the syntax in Indic languages. We include the IndicBERT \citep{kakwani-etal-2020-indicnlpsuite} and  MuRIL \citep{muril-2021}\footnote{Collectively called as ``Indic language models" here on}  models in our experiments, which are trained on Indic languages and English exclusively. We probe the models for four syntax-related tasks: PoS Tagging (POS), Syntax Tree-depth Prediction (STDP), Grammatical Case Marking (GCM), and Subject-Verb Agreement (SVA). In an attempt to capture the linguistic diversity in the Indian subcontinent, we include one language each from the Indo-Aryan (Hindi) and Dravidian (Tamil) language families in the dataset. In order to incorporate the prevalent phenomena of code-switching in these languages, we also include English-based script-mixed versions of these languages in our experiments. Our layer-wise probing experiments show that the Indic language models do not show syntactic localization and do not capture the syntax in Indic languages as efficiently as the other ``highly multilingual language models".\footnote{mBERT, DistilmBERT, and XLM-R}

\section{Background}
\label{section-background}
Language models are usually evaluated with information theory-based perplexity measures. While these perplexity measures might show good agreement with a language model's natural language understanding (NLU) capabilities, they do not capture language models' syntactic abilities efficiently \citep{tran-etal-2018-importance}. Evaluating language models' syntactic abilities is quite important in furthering the research towards developing human-like robust language models \citep{linzen-2020-accelerate}. Recent work has focused on the targeted syntactic evaluation of language models \citep{linzen-etal-2016-assessing, https://doi.org/10.1111/cogs.12414, gulordava-etal-2018-colorless, marvin-linzen-2018-targeted, mccoy-etal-2019-right, futrell-etal-2019-neural}, which takes inspiration from various psycholinguistic generalizations found in humans and assesses the role of syntax in the models' ability to perform various NLU tasks. While there has been significant progress towards building NLU evaluation benchmarks  \citep{wang-etal-2018-glue, DBLP:journals/corr/abs-1905-00537}, the work in developing syntactic evaluation benchmarks has been quite recent and limited \citep{gauthier-etal-2020-syntaxgym, hu-etal-2020-systematic, mueller-etal-2020-cross}.

Probing is an alternative paradigm that can be used for syntactic evaluation of language models. Probing deals with quantifying the amount of linguistic information encoded in the pre-trained representations of language models \citep{DBLP:conf/iclr/AdiKBLG17, conneau-etal-2018-cram}. While this does not provide a very detailed analysis of models' syntactic behavior, it can be used to compare the amount of syntactic information captured by the models, as well as the inner dynamics of how and where this information is encoded in the models \citep{hewitt-manning-2019-structural, jawahar-etal-2019-bert, lin-etal-2019-open, liu-etal-2019-linguistic, tenney2018what, rogers-bertology}. Probing methods are also relatively compute-efficient, as they do not involve training or fine-tuning the language models.

With the recent development in multilingual language modeling, numerous evaluation studies have been performed to test the models' multilingual and cross-lingual abilities \citep{DBLP:journals/corr/abs-1911-03310, ronnqvist-etal-2019-multilingual, pires-etal-2019-multilingual, wu-dredze-2019-beto, artetxe-etal-2020-cross}. Similar to monolingual language modeling paradigms, there has been significant work towards building multilingual evaluation benchmarks \citep{pmlr-v119-hu20b, kakwani-etal-2020-indicnlpsuite, liang-etal-2020-xglue}. However, these benchmarks do not cover syntactic evaluation efficiently, with POS tagging being the sole syntax-related task. Some recent studies have tried to address this by building multilingual and cross-lingual syntactic evaluation suites with subject-verb agreement tasks \citep{gulordava-etal-2018-colorless, mueller-etal-2020-cross}. 

Alongside the developments in multilingual NLP, there has been some recent progress towards advancing Indic NLP \citep{arora-2020-inltk, jain2020indictransformers, kakwani-etal-2020-indicnlpsuite, kunchukuttan2020indicnlp}. While there are numerous datasets available to benchmark and compare multilingual language models on NLU tasks in Indic languages, there is no such resource available for syntactic evaluation in Indic languages.\footnote{To the best of our knowledge} Most recent works \citep{jain2020indictransformers, kakwani-etal-2020-indicnlpsuite, muril-2021} rely on POS tagging with Universal Dependencies treebanks as the sole task to compare the syntactic abilities of Indic language models.

\begin{figure}[thbp]
  \includegraphics[width=\columnwidth, left]{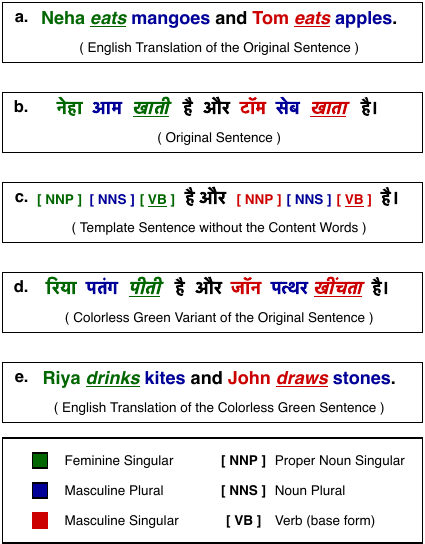}
  \caption{(a,b) - Comparing the linear word-order in English and Hindi; (b,c,d,e) - Constructing Colorless Green sentences in Hindi.}
  \label{fig-cgi}
\end{figure}

\section{Vyākarana: A Colorless Green Benchmark\footnote{Example sentences from the benchmark dataset can be found in Appendix~\ref{sec:data-appendix}}}
\label{section-vyākarana}
Syntax in Indic languages has specific characteristics which make it more challenging to model. Unlike English, most of the Indic languages follow a free linear word-order, with the default order being SOV (subject-object-verb). Hence, the linear distance (in terms of the number of intervening words) between the verb and its subject is usually larger in Indic languages as compared to English (Figure~\ref{fig-cgi}a; Figure~\ref{fig-cgi}b). Indic languages like Hindi and Tamil also have grammatical genders and grammatical number, where the words are morphologically inflected in agreement with the gender and count of their corresponding head nouns (Figure~\ref{fig-cgi}b). 


Usually, the syntactic evaluation of language models is conducted with syntax-related probing tasks \citep{jawahar-etal-2019-bert}, or with targeted syntactic evaluation through controlled psycholinguistic experiments \citep{gauthier-etal-2020-syntaxgym}. Performing such syntactic evaluations under a colorless green setting ensures that the models do not leverage any semantic or lexical cues and biases from the text to process its syntactic structure \citep{gulordava-etal-2018-colorless, DBLP:journals/corr/abs-1901-05287}. Most of the evaluation tasks from both of these categories can be designed with a syntactic dependencies treebank. Hence, we construct a colorless green treebank for Indic languages, which can be used to perform various syntactic evaluation tasks under a colorless green setting. 

\subsection{CG-HDTB: A Colorless Green Treebank for Hindi\footnote{In this section we use Hindi as a demonstrative language, where a similar approach is used for Tamil language.}}

We use the Hindi Universal Dependency Treebank (HDTB) \citep{bhathindi, palmer2009hindi} to construct a new Colorless Green treebank for Hindi: \emph{CG-HDTB}.\footnote{Similarly, we use the Tamil Universal Dependency Treebank (TTB) \citep{ramasamy-zabokrtsky-2012-prague} to construct a new Colorless Green treebank for Tamil: \emph{CG-TTB}.} We follow the method proposed by \citet{gulordava-etal-2018-colorless} to generate colorless green sentences. Given a sentence from the original HDTB treebank, we first convert it to a template colorless green sentence (Figure~\ref{fig-cgi}c) by removing all the content words (while retaining all the function words) from it. Each of the resulting gaps created by the removal of content words in the template sentence is then populated with a content word from another sentence in the treebank, where the grammatical information of the original and replacing content word is the same (Figure~\ref{fig-cgi}d). The resulting colorless green sentence still retains the same grammatical information and syntactic structure as the original sentence, whereas the random substitution of the content words ensures that it is rendered nonsensical (Figure~\ref{fig-cgi}a; Figure~\ref{fig-cgi}e).

\begin{table}[htbp]
\centering
\resizebox{0.80\columnwidth}{!}{
\begin{tabular}{@{}ccccc@{}}
\toprule
\textbf{PoS Tag}   & \textbf{Gen} & \textbf{Num} & \textbf{Case} & \textbf{Per} \\ \midrule
\textbf{Noun}      & \checkmark   & \checkmark   & \checkmark    &              \\
\textbf{Verb}      & \checkmark   & \checkmark   &               & \checkmark   \\
\textbf{Adjective} & \checkmark   & \checkmark   & \checkmark    &              \\
\textbf{Adverb}    & \checkmark   & \checkmark   & \checkmark    &              \\ \bottomrule
\end{tabular}
}
\caption{The grammatical information taken into account for various PoS categories of content words \citep{vikram2013morphology}.}
\label{tab:data-grammar-info}
\end{table}

\begin{table*}[htbp]
\centering
\resizebox{\textwidth}{!}{
\begin{tabular}{@{}ccccccc@{}}
\toprule
\multirow{2}{*}{\textbf{Data Split}} & \multicolumn{3}{c}{\textbf{HDTB}}                           & \multicolumn{3}{c}{\textbf{CG-HDTB}}                        \\ \cmidrule(l){2-4} \cmidrule(l){5-7}
                                     & \textbf{Sentences} & \textbf{Masculine} & \textbf{Feminine} & \textbf{Sentences} & \textbf{Masculine} & \textbf{Feminine} \\ \cmidrule(l){1-1} \cmidrule(l){2-4} \cmidrule(l){5-7}
\textbf{Train} & 13304 & 104389 (70.37\%) & 43951 (29.63\%) & 6736  & 39178 (51.72\%)  & 36578 (48.28\%)  \\
\textbf{Dev}   & 1659  & 13116 (70.47\%)  & 5496 (29.53\%)  & 6628  & 38638 (51.90\%)  & 35810 (48.10\%)  \\
\textbf{Test}  & 1684  & 13253 (69.98\%)  & 5686 (30.02\%)  & 53124 & 308034 (51.91\%) & 285322 (48.09\%) \\ \bottomrule
\end{tabular}
}
\caption{The number of sentences and the statistics for token-level grammatical gender feature in the HDTB and CG-HDTB treebank (the statistics are same for CG-HDTB and csCG-HDTB treebanks).}
\label{tab:treebank-stats}
\end{table*}

The grammatical information taken into consideration while substituting the content words includes the Grammatical Gender \textbf{(Gen)} (masculine/feminine), Grammatical Number \textbf{(Num)} (singular/plural), Grammatical Case \textbf{(Case)} (Section~\ref{sec:gcm}), and Person \textbf{(Per)} (first/second/third). This makes sure that the replacing content word is consistent with syntax and the morphological inflections of the original content word, keeping the morphosyntax of the sentence intact. The grammatical information required for every PoS category is different, as shown in Table~\ref{tab:data-grammar-info}. The HDTB treebank has an imbalanced count of Grammatical Gender features. Following an uncontrolled and random substitution of content words can result in a gender-imbalanced dataset of colorless green sentences. We eliminate this imbalance by controlling the grammatical gender information in the colorless green sentences. For every sentence in the HDTB treebank, we generate four colorless green sentences, where the gender encoding of the sentences is:

\begin{compactenum}
   \item Exactly \textbf{same} as that of the original sentence.
   \item Exactly \textbf{opposite} as that of the original sentence.
   \item Entirely \textbf{masculine}.
   \item Entirely \textbf{feminine}.
\end{compactenum}

Apart from the PoS categories of content words mentioned in Table~\ref{tab:data-grammar-info}, we also adjust the gender-inflected Adposition tokens in the colorless green sentences in order to maintain the morphosyntax of the sentence. The statistics for the Grammatical Gender feature in the gender-imbalanced HDTB treebank and the gender-balanced CG-HDTB treebank are shown in Table~\ref{tab:treebank-stats}. Further, we swap the train and test set of the HDTB treebank while generating colorless green sentences (Table~\ref{tab:treebank-stats}). This allows the test set of CG-HDTB to be significantly bigger than the training set, providing a challenging test setting for the models' syntactic generalization capabilities.

Unlike Hindi, there is no publicly available treebank for Hindi-English code-switched setting with all the features that are required to obtain the colorless green sentences. Hence, we artificially create a code-switched treebank for Hindi-English: \textbf{csCG-HDTB}. The csCG-HDTB treebank is a parallel code-switched version of the CG-HDTB treebank. Following \citet{muril-2021}'s approach, we use the Indic-Trans library\footnote{\url{https://github.com/libindic/indic-trans}} to transliterate the tokens (Hindi-Devanagari script $\rightarrow$ English-Latin script) from each sentence in the CG-HDTB treebank.\footnote{Similarly, we create a code-switched treebank for Tamil-English: \textbf{csCG-TTB}.} While this only helps in incorporating the script-mixing aspect of code-switching, the approach can be scaled easily across different datasets and languages, eliminating the need of explicit annotations of morphosyntactic features on code-switched data.

\subsection{Evaluation Tasks}

\begin{table*}[thbp]
\centering
\resizebox{0.8\textwidth}{!}{
\begin{tabular}{@{}ccccccccc@{}}
\toprule
\multirow{2}{*}{\textbf{Data Split}} & \multicolumn{4}{c}{\textbf{HDTB}}                          & \multicolumn{4}{c}{\textbf{CG-HDTB}}                       \\ \cmidrule(l){2-5} \cmidrule(l){6-9}
                                     & \textbf{POS} & \textbf{STDP} & \textbf{GCM} & \textbf{SVA} & \textbf{POS} & \textbf{STDP} & \textbf{GCM} & \textbf{SVA} \\ \cmidrule(l){1-1} \cmidrule(l){2-5} \cmidrule(l){6-9}
\textbf{Train} & 281057 & 13304 & 151275 & 17034 & 141720  & 6736  & 76836  & 8536  \\
\textbf{Dev}   & 35430  & 1659  & 19209  & 2127  & 140844 & 6628  & 76108 & 8508  \\
\textbf{Test}  & 35217  & 1684  & 19027  & 2134  & 1123964  & 53124 & 605068  & 68136 \\ \bottomrule
\end{tabular}
}
\caption{The number of tokens (for POS and GCM tasks) and the number of sentences (for STDP and SVA tasks) in the HDTB and CG-HDTB treebank (the statistics are same for CG-HDTB and csCG-HDTB treebanks).}
\label{tab:probing-stats}
\end{table*}

The benchmark comprises four syntactic evaluation tasks. We borrow certain design principles for the benchmark from the XTREME benchmark \citet{pmlr-v119-hu20b}:

\begin{compactenum}
   \item \textbf{Task Difficulty:} The colorless-green setting, and a sufficiently large test-set ensure a certain level of difficulty across all the tasks. Further, some tasks require capturing long-range relationships and morphological inflections, which make them more challenging.
   \item \textbf{Task Diversity:} We include both the token-level and the sentence-level evaluation tasks in the benchmark. This ensures that the benchmark evaluates representations from both the granularities.
   \item \textbf{Data and Training Efficiency:} Under the colorless-green setting, each task has a large test-train ratio in terms of data samples (Table~\ref{tab:probing-stats}). Hence, the datasets for the evaluation tasks are quite challenging in terms of data and training efficiency.
\end{compactenum}

We use the HDTB, CG-HDTB, and csCG-HDTB treebanks to construct datasets for all the evaluation tasks.\footnote{Similarly, we use the CG-TTB and csCG-TTB treebanks to conduct the experiments with the Tamil language (Appendix~\ref{sec:tamil-appendix}).}  The statistics for the constructed datasets are given in Table~\ref{tab:probing-stats}.

\paragraph{PoS Tagging (POS):}  While PoS tagging is a very weak construct for a syntactic evaluation task, it is a preliminary step in syntactic processing. It is also the only existing syntax-related evaluation task used to compare currently available Indic language models. Hence, we include PoS tagging as the first evaluation task in the benchmark. The task is designed under a token-level multi-class single-label classification setting. We use the UPOS tags from the treebanks as the ground-truth labels for this task. 

\paragraph{Syntax Tree-depth Prediction (STDP):} Following the work done by \citet{conneau-etal-2018-cram} and \citet{jawahar-etal-2019-bert}, we use the dependency trees from the treebanks to perform syntax tree-depth prediction task. Successful prediction of the depth of a dependency tree depicts the model's ability to get a surface-level estimate of a given sentence's syntactic structure. The task is designed under a sentence-level multi-class single-label classification setting.

\paragraph{Grammatical Case Marking (GCM):} 
\label{sec:gcm}
Given the nature of morphosyntax in Indic languages, case marking is an essential syntactic evaluation task. Unlike languages like English, a fixed linear word-order cannot be used to perform case marking in Indic languages. Instead, one must rely on morphological inflections and Adposition tokens in the sentence to assign appropriate grammatical cases. Moreover, a single token can be marked with a combination of multiple cases. In the HDTB treebank, the grammatical case feature has the following seven unique values: \textbf{accusative}, \textbf{nominative}, \textbf{accusative-inessive}, \textbf{dative-accusative}, \textbf{ergative-accusative}, \textbf{genitive-accusative}, and \textbf{instrumental-accusative}. Hence, this task is designed under a token-level multi-class single-label classification setting.

\paragraph{Subject-Verb Agreement (SVA):} We include subject-verb agreement as an evaluation task to study the long-range syntactic dependencies in Indic languages. Unlike English, the subject-verb agreement in Indic languages is dependent on the grammatical number as well as the grammatical gender of the tokens. Following the work done by \citet{linzen-etal-2016-assessing}, we include all the tokens preceding (and excluding) the target verb in the sentence. Given such a sentence, the task is to predict the target verb's grammatical count and grammatical gender in agreement with its subject (head noun). Given that most of the Indic languages follow a free linear word-order and the default order is SOV (subject-object-verb), the number of intervening nouns is significantly higher than in English. Moreover, there is a high probability of the object acting as an attractor noun (Figure~\ref{fig-cgi}). Hence, the SVA task is significantly more challenging in Indic languages. The task is designed under a sentence-level multi-class single-label classification setting, given the four possible ground truth labels: \textbf{masculine-singular}, \textbf{masculine-plural}, \textbf{feminine-singular}, and \textbf{feminine-plural}.

\begin{table*}[htbp]
\centering
\resizebox{\textwidth}{!}{
\begin{tabular}{@{}cccccccccccc@{}}
\toprule
\multirow{2}{*}{\textbf{Treebank}} &
  \multirow{2}{*}{\textbf{Task}} &
  \multicolumn{5}{c}{\textbf{Last Layer}} &
  \multicolumn{5}{c}{\textbf{Best Layer}} \\  \cmidrule(l){3-7} \cmidrule(l){8-12}
 &
   &
  \textbf{mBERT} &
  \textbf{XLM-R} &
  \textbf{DistilmBERT} &
  \textbf{IndicBERT} &
  \textbf{MuRIL} &
  \textbf{mBERT} &
  \textbf{XLM-R} &
  \textbf{DistilmBERT} &
  \textbf{IndicBERT} &
  \textbf{MuRIL} \\ \cmidrule(l){1-2} \cmidrule(l){3-7} \cmidrule(l){8-12}
\multirow{5}{*}{\textbf{HDTB}} &
  \textbf{POS} &
  0.9332 &
  \textbf{0.9567} &
  0.8860 &
  0.7827 &
  0.7290 &
  0.9409 (8) &
  \textbf{0.9615 (8)} &
  0.8955 (5) &
  0.8105 (2) &
  0.7728 (1) \\
 &
  \textbf{STDP} &
  \textbf{0.3886} &
  0.1848 &
  0.3815 &
  0.3825 &
  0.2588 &
  0.4138 (3) &
  0.3657 (1) &
  \textbf{0.4247 (3)} &
  0.3995 (3) &
  0.3969 (8) \\
 &
  \textbf{GCM} &
  0.7501 &
  \textbf{0.7733} &
  0.7345 &
  0.6509 &
  0.6350 &
  0.7775 (8) &
  \textbf{0.8006 (8)} &
  0.7495 (4) &
  0.6847 (3) &
  0.6522 (1) \\
 &
  \textbf{SVA} &
  \textbf{0.7083} &
  0.6208 &
  0.7015 &
  0.5122 &
  0.5178 &
  \textbf{0.7083 (12)} &
  0.6208 (12) &
  0.7015 (6) &
  0.5125 (3) &
  0.5718 (12) \\ \cmidrule(l){2-2} \cmidrule(l){3-7} \cmidrule(l){8-12}
 &
  \textbf{Average} &
  \textbf{0.6951} &
  0.6339 &
  0.6759 &
  0.5821 &
  0.5352 &
  \textbf{0.7101} &
  0.6872 &
  0.6928 &
  0.6018 & 
  0.5984 \\ \cmidrule(l){1-2} \cmidrule(l){3-7} \cmidrule(l){8-12}
\multirow{5}{*}{\textbf{CG-HDTB}} &
  \textbf{POS} &
  0.8894 &
  \textbf{0.9143} &
  0.8448 &
  0.7418 &
  0.6807 &
  0.8932 (11) &
  \textbf{0.9232 (8)} &
  0.8543 (5) &
  0.7642 (6) &
  0.7030 (2) \\
 &
  \textbf{STDP} &
  0.3455 &
  0.2138 &
  0.3254 &
  \textbf{0.3566} &
  0.2318 &
  0.3730 (6) &
  0.3232 (3) &
  \textbf{0.3881 (5)} &
  0.3756 (4) &
  0.3570 (10) \\
 &
  \textbf{GCM} &
  0.6886 &
  \textbf{0.6967} &
  0.6733 &
  0.6216 &
  0.5463 &
  0.7050 (7) &
  \textbf{0.7234 (8)} &
  0.6968 (4) &
  0.6479 (2) &
  0.5878 (3) \\
 &
  \textbf{SVA} &
  0.6003 &
  0.5935 &
  \textbf{0.6040} &
  0.3990 &
  0.5568 &
  \textbf{0.6140 (8)} &
  0.5935 (12) &
  0.6040 (6) &
  0.4451 (2) &
  0.5568 (12) \\ \cmidrule(l){2-2} \cmidrule(l){3-7} \cmidrule(l){8-12}
 &
  \textbf{Average} &
  \textbf{0.6310} &
  0.6046 &
  0.6119 &
  0.5298 &
  0.5039 &
  \textbf{0.6463} &
  0.6408 &
  0.6358 &
  0.5582 &
  0.5511 \\ \cmidrule(l){1-2} \cmidrule(l){3-7} \cmidrule(l){8-12}
\multirow{5}{*}{\textbf{csCG-HDTB}} &
  \textbf{POS} &
  0.7543 &
  \textbf{0.7881} &
  0.7422 &
  0.7303 &
  0.6752 &
  0.7744 (7) &
  \textbf{0.8079 (10)} &
  0.7596 (5) &
  0.7608 (2) &
  0.7011 (1) \\
 &
  \textbf{STDP} &
  0.3104 &
  0.3429 &
  0.3531 &
  \textbf{0.3805} &
  0.2457 &
  0.3879 (7) &
  0.3584 (3) &
  0.3843 (5) &
  \textbf{0.3909 (3)} &
  0.3676 (7) \\
 &
  \textbf{GCM} &
  0.6415 &
  0.6456 &
  \textbf{0.6533} &
  0.6306 &
  0.5653 &
  0.6604 (8) &
  \textbf{0.6683 (8)} &
  0.6549 (5) &
  0.6535 (8) &
  0.5859 (3) \\
 &
  \textbf{SVA} &
  0.5334 &
  \textbf{0.5629} &
  0.5255 &
  0.5389 &
  0.5262 &
  0.5650 (1) &
  0.5629 (12) &
  \textbf{0.5752 (3)} &
  0.5389 (12) &
  0.5647 (11) \\ \cmidrule(l){2-2} \cmidrule(l){3-7} \cmidrule(l){8-12}
 &
  \textbf{Average} &
  0.5599 &
  \textbf{0.5849} &
  0.5685 &
  0.5701 &
  0.5031 &
  0.5969 &
  \textbf{0.5994} &
  0.5935 &
  0.5860 &
  0.5548 \\ \bottomrule 
\end{tabular}
}
\caption{The weighted-F1 scores for the last layer and the best layer (layer-number mentioned in parenthesis) for the layer-wise probing experiments with the Hindi language data. (Corresponding results for the Tamil language data can be found in Appendix~\ref{sec:appendix-tamil-results})}
\label{tab:results-last-best}
\end{table*}

\section{Experiments\footnote{In this section we discuss the experiments with the Hindi language data. Corresponding details about the experiments with the Tamil language data can be found in Appendix~\ref{sec:tamil-appendix}.}}
\label{section-experiments}

\subsection{Experimental Setup}
We use the datasets constructed for the evaluation tasks to perform layer-wise probing\footnote{We do not fine-tune the models due to computational limitations. Instead, we report the probing metrics of the last layer and the best-performing layer of every model (Table~\ref{tab:results-last-best}).} of token-level and sentence-level representations from the five transformer-based multilingual language models: mBERT \citep{devlin-etal-2019-bert}, XLM-R \citep{conneau-etal-2020-unsupervised}, DistilmBERT \citep{DBLP:journals/corr/abs-1910-01108}, IndicBERT \citep{kakwani-etal-2020-indicnlpsuite}, and MuRIL \citep{muril-2021}. We use a single linear-layer (initialized with the same weights across all the experiments) as the probing classifier for all the tasks and models. Since POS and GCM are token-level tasks, we use the first-subword token-embeddings as the input to the probing classifier. Whereas we use the special sentence-token embeddings for the sentence-level STDP and SVA tasks. We use Hugging Face's \emph{transformers} library \citep{wolf-etal-2020-transformers} to access the pre-trained instances of these language models. The probing classifier is trained using the PyTorch library \citep{NEURIPS2019_bdbca288}. Given that all the evaluation tasks are designed under a multi-class single-label classification setting, we monitor the weighted-F1 scores to evaluate the models (Table~\ref{tab:results-last-best}).

\subsection{Probing Results}
\label{sec:probing}
Overall, due to the lack of semantic and lexical cues, and a larger test-set, all the models find the CG-HDTB data more challenging than the HDTB data (Table~\ref{tab:results-last-best}). The performance across all the models and tasks further deteriorates with the code-switched csCG-HDTB data. Given the considerably larger scale (model and pre-training data) of the XLM-R model, for the token-level tasks (POS and GCM), it significantly outperforms the other models for all the three treebanks.  Whereas, for the sentence-level tasks (STDP and SVA), mBERT and DistilmBERT perform better than the rest of the models for HDTB and CG-HDTB data. This might be explained by the sentence-level pre-training tasks (Next Sentence Prediction) used by mBERT and DistilmBERT. Overall, for Hindi data, mBERT shows the best average weighted-F1 scores across the four tasks, and the Indic models perform the worst. On the other hand, XLM-R significantly outperforms the rest of the models under a code-switched Hindi-English setting (Table~\ref{tab:results-last-best}).

Even though the IndicBERT and MuRIL models outperform mBERT and XLM-R on many downstream NLU tasks \citep{kakwani-etal-2020-indicnlpsuite, muril-2021}, they consistently fail to catch-up in syntactic evaluation. Even DistilmBERT, a relatively smaller model, outperforms IndicBERT and MuRIL across all the tasks. There might be multiple plausible reasons behind this finding. Both IndicBERT, and MuRIL perform masked word/token language modeling and do not have a sentence-level pre-training task. Hence, while they are outperformed by mBERT and DistilmBERT, they outperform XLM-R on most of the sentence-level tasks with Hindi data. Moreover, they are significantly smaller in architecture size (IndicBERT) and dataset size as compared to the other models. mBERT, XLM-R, and DistilmBERT are highly multilingual language models, pre-trained on more than 100 languages. This might provide them with the linguistic and typological generalization required for modeling morphosyntax more efficiently than Indic models, which are only trained on a handful of Indic languages and English. Even though the MuRIL model is pre-trained with artificially generated parallel translated and transliterated sentences in Indic languages and their English counterparts, it is outperformed by XLM-R on the csCG-HDTB Hindi-English code-switched data. The XLM-R model has a small amount of natural code-switched data in its pre-training corpus, which might be the reason behind its dominance under the code-switching setting.

\begin{figure*}
  \begin{subfigure}[t]{\columnwidth}
    \centering
    \includegraphics[width=\columnwidth]{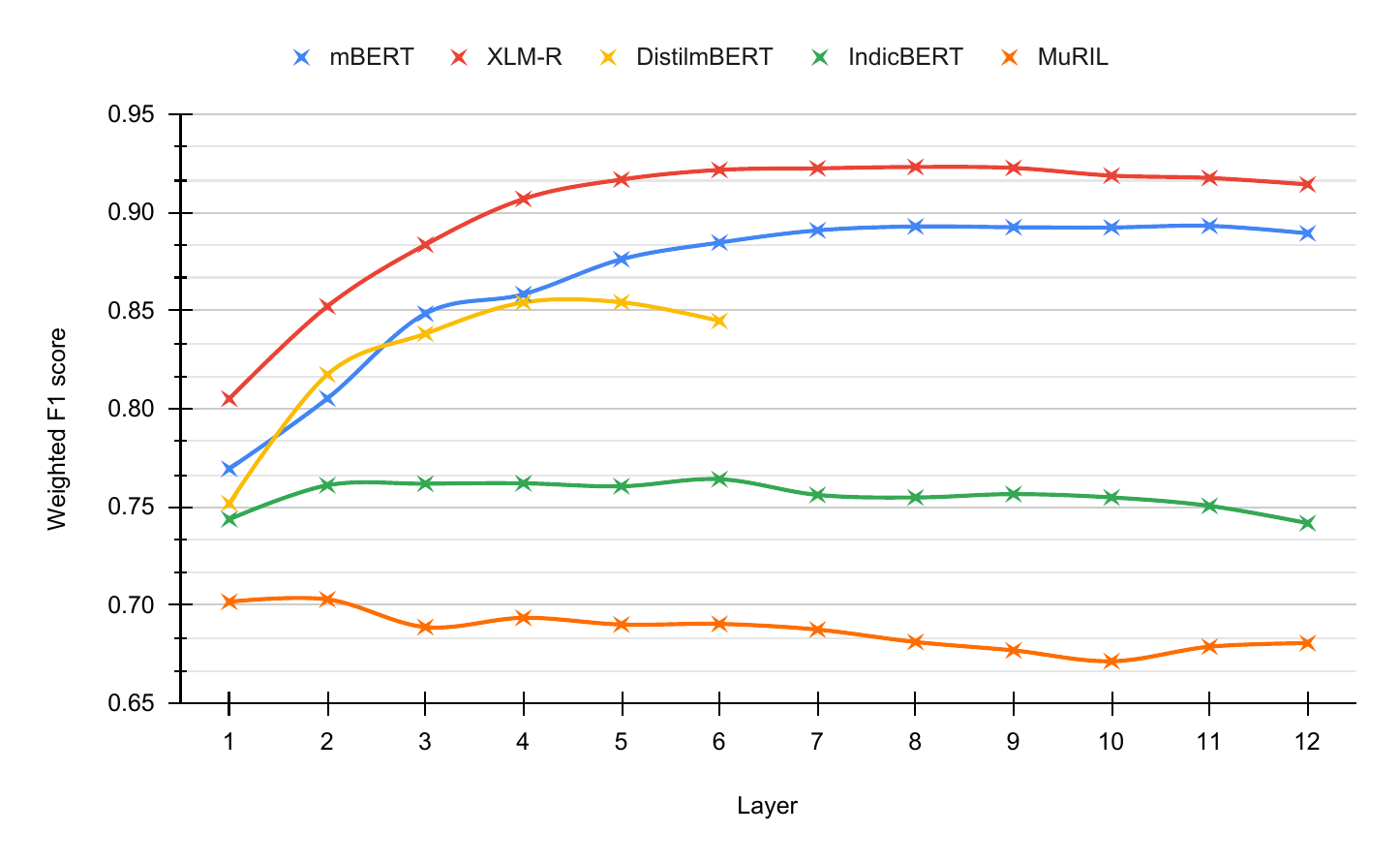}
    \caption{PoS Tagging (POS)}
    \label{fig-cgi-hindi-a}
  \end{subfigure}
  \hfill
  \begin{subfigure}[t]{\columnwidth}
    \centering
    \includegraphics[width=\columnwidth]{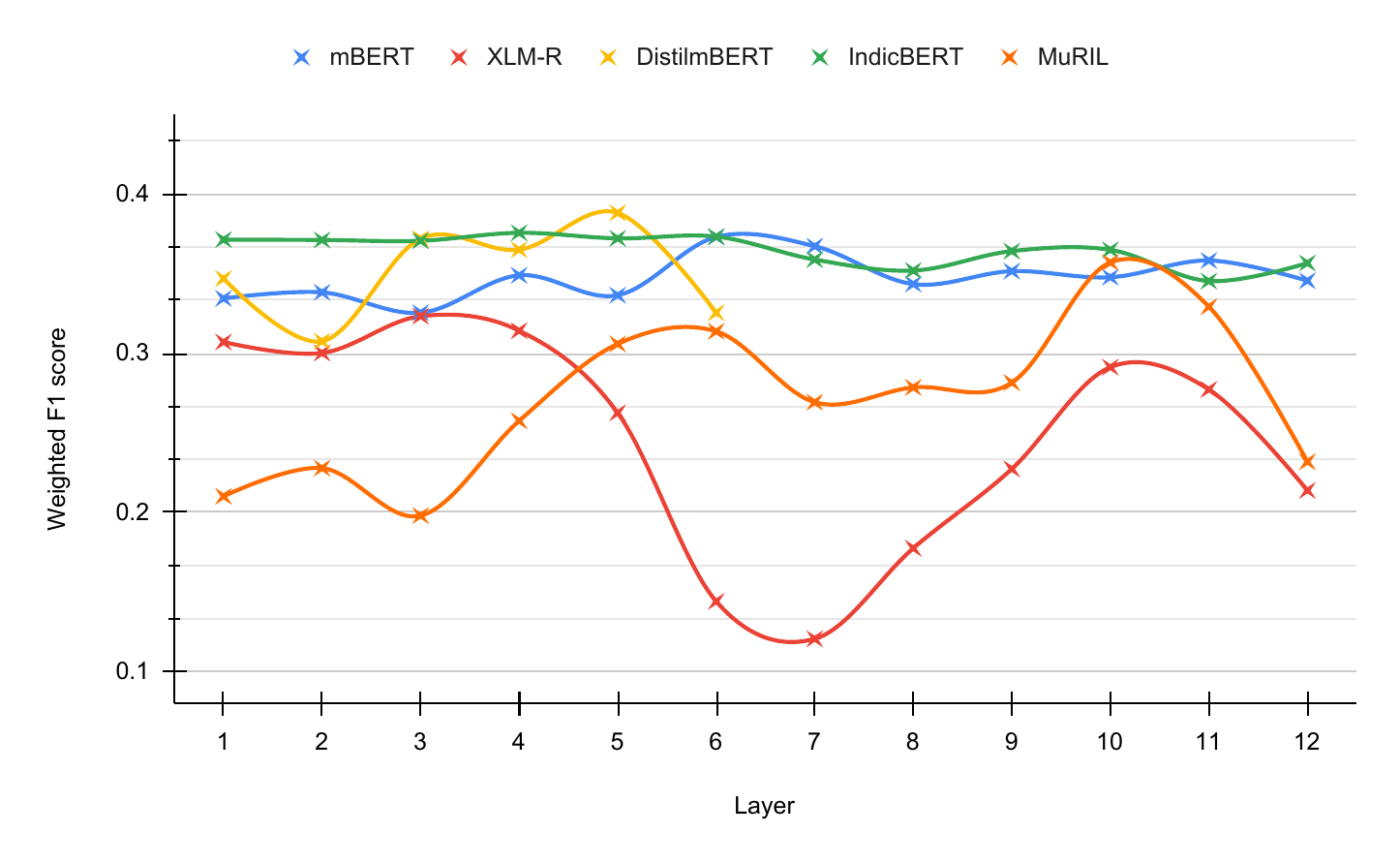}
    \caption{Syntax Tree-depth prediction (STDP)}
    \label{fig-cgi-hindi-b}
  \end{subfigure}

  \medskip

  \begin{subfigure}[t]{\columnwidth}
    \centering
    \includegraphics[width=\columnwidth]{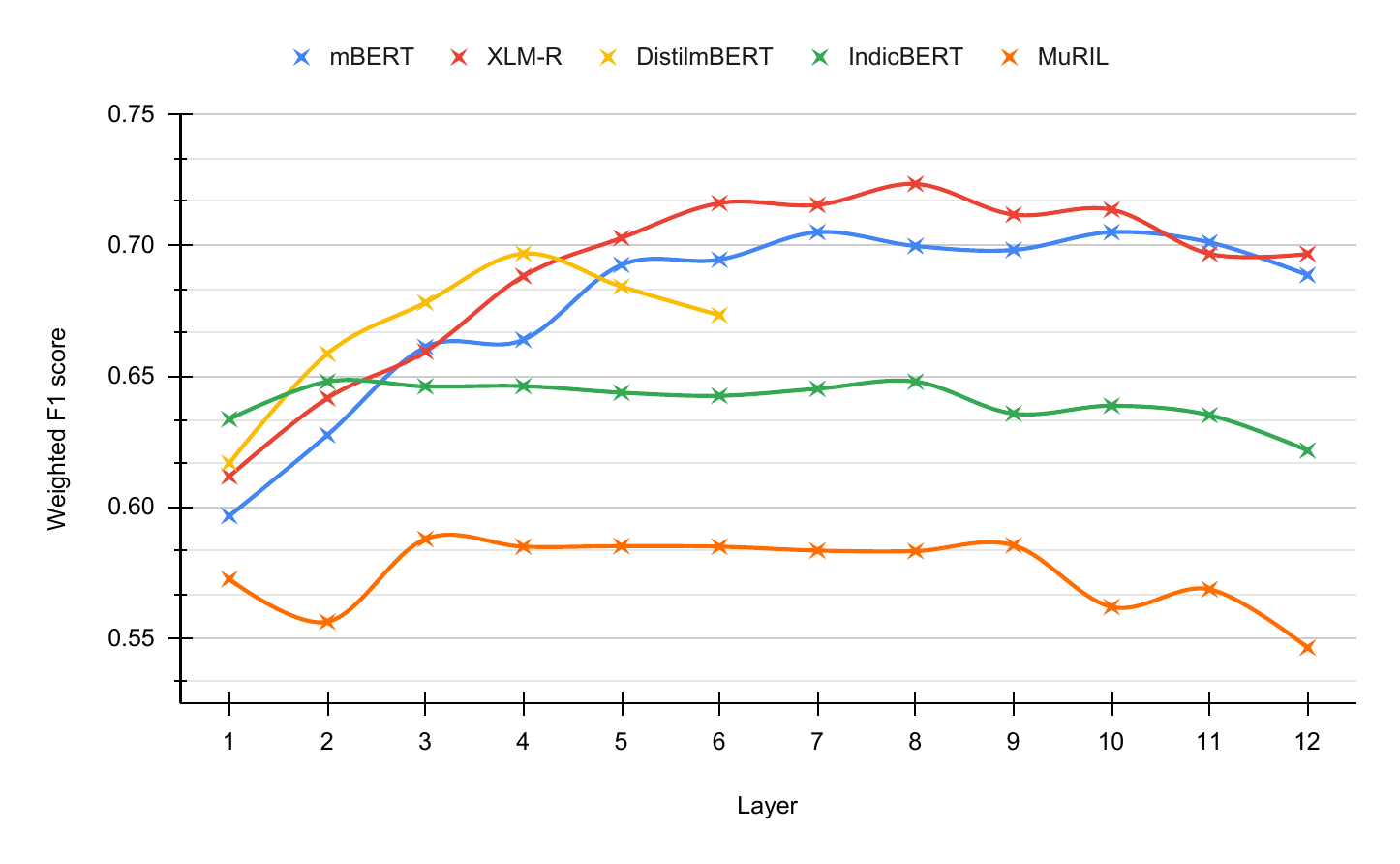}
    \caption{Grammatical Case Marking (GCM)}
    \label{fig-cgi-hindi-c}
  \end{subfigure}
  \hfill
  \begin{subfigure}[t]{\columnwidth}
    \centering
    \includegraphics[width=\columnwidth]{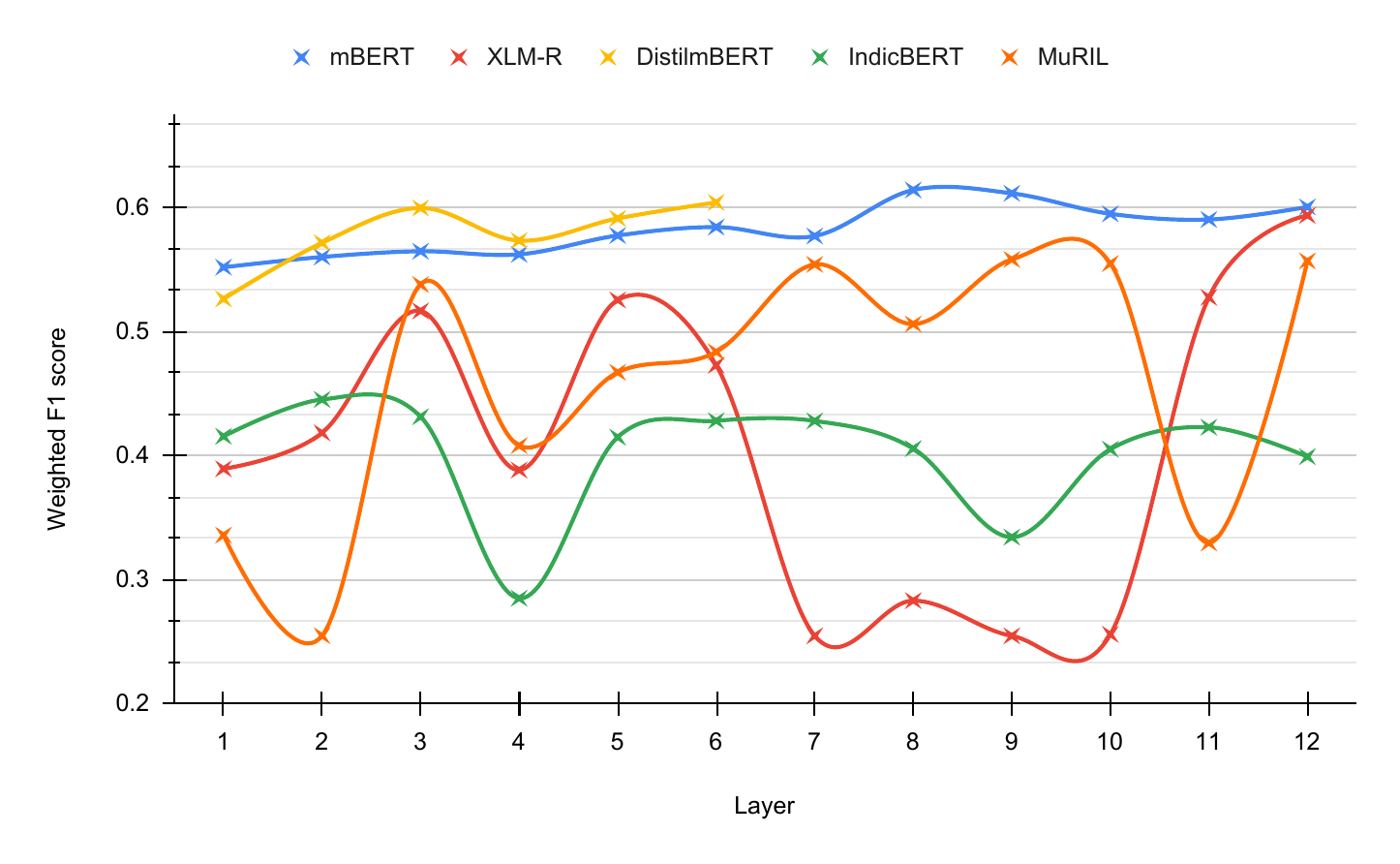}
    \caption{Subject-Verb Agreement (SVA)}
    \label{fig-cgi-hindi-d}
  \end{subfigure}
  \caption{Weighted F1 scores for the layer-wise probing experiments with the CG-HDTB test set for the \textcolor{blue}{mBERT}, \textcolor{yellow}{DistilmBERT}, \textcolor{red}{XLM-R}, \textcolor{green}{IndicBERT}, and \textcolor{orange}{MuRIL} models.}
  \label{fig-cgi-hindi-results}
\end{figure*}

\subsection{Localization of Syntactic Knowledge}
\label{sec:localization}
It has been previously shown that linguistic knowledge is localized in the monolingual BERT \citep{rogers-bertology}. While surface-level linguistic information like linear-word order and sentence-length is captured in the lower layers of the model \citep{jawahar-etal-2019-bert, lin-etal-2019-open}, the syntactic knowledge is found in the middle-layers of the model \citep{hewitt-manning-2019-structural, jawahar-etal-2019-bert}. While basic syntactic operations are encoded in relatively lower-layers \citep{jawahar-etal-2019-bert, tenney-etal-2019-bert}, more complex tasks with long-range syntactic dependencies are performed best by relatively higher layers \citep{DBLP:journals/corr/abs-1901-05287}. \citet{rogers-bertology} explain this localization of syntactic knowledge using \citet{liu-etal-2019-linguistic}'s finding that the middle layers of BERT are most generalizing across tasks and the deeper layers are more task-specific. We investigate the localization of syntactic knowledge in multilingual language models by plotting the layer-wise performance across the various evaluation tasks with CG-HDTB data (Figure~\ref{fig-cgi-hindi-results}).

For token-level tasks (POS and GCM) (Figure~\ref{fig-cgi-hindi-a}~\ref{fig-cgi-hindi-c}), we find a pattern similar to that of monolingual BERT in mBERT, DistilmBERT, and XLM-R, where the weighted-F1 scores peak at the middle layers. The same, however, is not true for IndicBERT and MuRIL, which show declining weighted-F1 scores with increasing depth in the model. This suggests that with increasing depth, the layers become more task-specific. This is an unusual behavior for the Indic models, where investigating the cause for this should be an interesting research direction. The XLM-R model consistently outperforms the rest of the models across all the layers. Whereas the Indic language models consistently fall behind the other highly multilingual models. Between the Indic models, IndicBERT significantly outperforms MuRIL across all the layers.

For sentence-level tasks (STDP and SVA) (Figure~\ref{fig-cgi-hindi-b}~\ref{fig-cgi-hindi-d}), XLM-R, IndicBERT and MuRIL show no clear localization pattern\footnote{(with the exception of IndicBERT on task STDP)}. mBERT and DistilmBERT show a consistent performance across all the layers, outperforming the rest of the models for both tasks. Similar to \citet{jawahar-etal-2019-bert}'s findings of the STDP task with monolingual BERT, they peak at the middle layers. However, the difference between the layers' performance is not very significant (Figure~\ref{fig-cgi-hindi-b}). IndicBERT performs particularly well on the STDP task, showing comparable metrics with mBERT and DistilmBERT. While MuRIL starts lowest for STDP task, it eventually manages to outperform XLM-R, but still consistently lags behind IndicBERT (Figure~\ref{fig-cgi-hindi-b}). Whereas, for SVA task, MuRIL performs better than IndicBERT for most of the layers (Figure~\ref{fig-cgi-hindi-d}). XLM-R model shows a dip in performance in the middle layers for both the tasks, which is exactly opposite to that of the token-level tasks and previous findings with monolingual BERT.

\section{Conclusion}
\label{section-conclusion}
This work presents a gender-balanced benchmark evaluation of Colorless Green sentences in Indic languages for syntactic testing of multilingual language models. By doing so, we aim to address the existing research gap in syntactic testing in Indic NLP. We introduce four new treebanks: CG-HDTB, csCG-HDTB, CG-TTB, and csCG-TTB for this purpose. We use the datasets from the treebanks to perform four syntax-related benchmarking evaluation tasks. In our experiments, we perform layer-wise probing of token-level and sentence-level representations from five different multilingual language models. Our experiments reveal that performing syntactic evaluation under a colorless green setting ensures that the semantic and lexical cues do not add evaluation artifacts in the form of higher-than-actual performance metrics. Further, the experiments also reveal that the multilingual language models suffer significantly under a code-switched setting, which is an important aspect in Indic languages. Overall, our experiments show that the currently available Indic language models do not capture syntax as efficiently as the other highly multilingual language models. Indic models seem to outperform other multilingual language models on various NLU tasks in Indic languages even with a certain lack of syntactic knowledge. This might suggest that the Indic language models do not rely significantly on syntax while making inferences on such NLU tasks. We also observe that training on large artificial corpora of transliterated texts do not help Indic language models in capturing syntax under a code-switched setting. 

Further, we find that unlike the other highly multilingual language models, the currently available Indic language models do not show any syntactic localization in the middle layers. This is a unique behavior to the multilingual Indic language models, the cause and effects for which can be investigated in another independent study. While the current work only covers the Hindi and Tamil languages, it lays down the framework for performing syntactic evaluation in other Indic languages. 

\section{Development and Accessibility}
\label{section-accessibility}
As a continuation of the current work, we aim to cover more Indic languages and their code-switched counterparts in the benchmark. We aim to cover typologically different languages from both the Indo-Aryan, and the Dravidian language families, with a special focus on low-resourced Indic languages. This can be an extremely challenging task. Developing datasets for syntactic evaluation requires a certain level of linguistic expertise, which is usually found in the trained and native speakers of that particular language. While the methods used in this work relax these requirements upto some extent, they are dependent on existing syntactic dependency annotations. Such fine-grained annotations are not readily available for a majority of the Indic languages. Hence, we plan to open-source the benchmark datasets. We aim to continue developing the benchmark with open-source contributions from the trained and native speakers of various Indic languages.

The treebank datasets used for this benchmark are publicly available for benchmarking and development purposes on GitHub and the Hugging Face Datasets platform \cite{lhoest2021datasets}.\footnote{\url{https://github.com/rajaswa/indic-syntax-evaluation}}

\section*{Acknowledgments}
We would like to thank Greg Durrett for his feedback on an early draft of this work. We would also like to thank K. A. Geetha for contributing to our discussions around the work for the Tamil language. 

\bibliographystyle{acl_natbib}
\bibliography{anthology, custom}

\newpage
\clearpage

\appendix

\section{Dataset Examples}
\label{sec:data-appendix}

\begin{center}{
    \includegraphics[width=\textwidth, left]{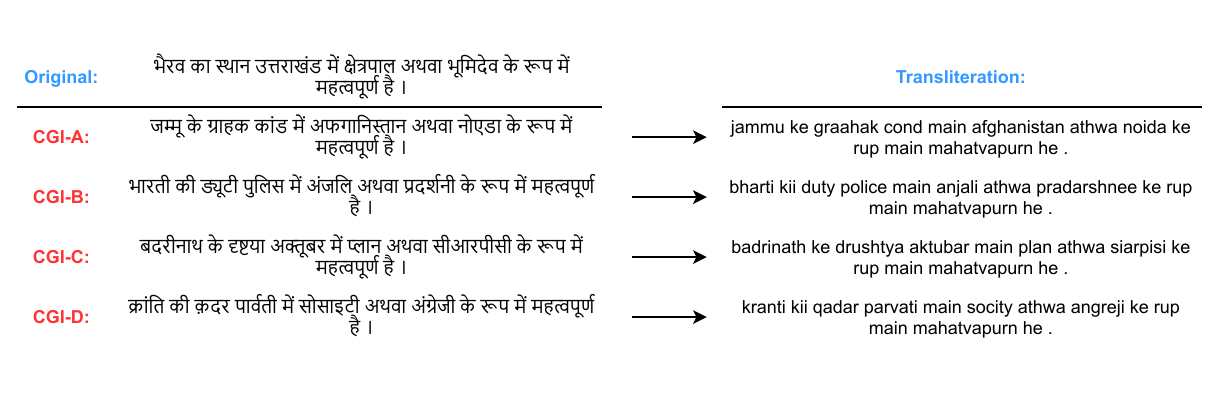}
    \captionof{figure}{Example sentences from the CG-HDTB (left) and csCG-HDTB (right) treebank corresponding to an original sentence from the HDTB (top-left) treebank.}}
    \label{fig-data-hindi}
\end{center}

\begin{center}{
    \includegraphics[width=\textwidth, left]{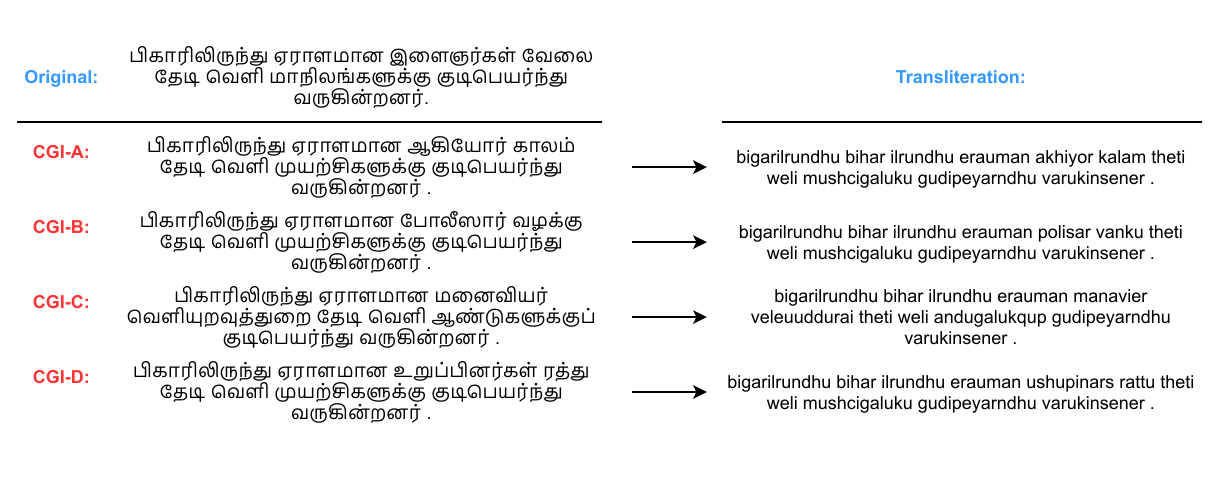}
    \captionof{figure}{Example sentences from the CG-TTB (left) and csCG-TTB (right) treebank corresponding to an original sentence from the TTB (top-left) treebank.}}
    \label{fig-data-tamil}
\end{center}




\FloatBarrier
\begin{table*}[htbp]
\centering
\resizebox{\textwidth}{!}{
\begin{tabular}{@{}cccccccccccc@{}}
\toprule
\multirow{2}{*}{\textbf{Treebank}} &
  \multirow{2}{*}{\textbf{Task}} &
  \multicolumn{5}{c}{\textbf{Last Layer}} &
  \multicolumn{5}{c}{\textbf{Best Layer}} \\  \cmidrule(l){3-7} \cmidrule(l){8-12}
 &
   &
  \textbf{mBERT} &
  \textbf{XLM-R} &
  \textbf{DistilmBERT} &
  \textbf{IndicBERT} &
  \textbf{MuRIL} &
  \textbf{mBERT} &
  \textbf{XLM-R} &
  \textbf{DistilmBERT} &
  \textbf{IndicBERT} &
  \textbf{MuRIL} \\ \cmidrule(l){1-2} \cmidrule(l){3-7} \cmidrule(l){8-12}
 \cmidrule(l){1-2} \cmidrule(l){3-7} \cmidrule(l){8-12}
\multirow{5}{*}{\textbf{CG-TTB}} &
  \textbf{POS} &
  \textbf{0.7444} &
  0.7336 &
  0.6431 &
  0.5874 &
  0.4539 &
  0.7719 (9) &
  \textbf{0.7946 (7)} &
  0.6809 (4) &
  0.6327 (6) &
  0.4741 (5) \\
 &
  \textbf{STDP} &
  0.0947 &
  0.0716 &
  \textbf{0.2315} &
  0.0716 &
  0.0898 &
  0.2209 (10) &
  0.0898 (6) &
  \textbf{0.2819 (3)} &
  0.0716 (1) &
  0.2051 (11) \\
 &
  \textbf{GCM} &
  \textbf{0.7319} &
  0.6800 &
  0.6336 &
  0.5864 &
  0.5878 &
  0.7966 (7) &
  \textbf{0.8187 (6)} &
  0.6765 (4) &
  0.5864 (12) &
  0.6040 (1) \\
  \cmidrule(l){2-2} \cmidrule(l){3-7} \cmidrule(l){8-12}
 &
  \textbf{Average} &
  \textbf{0.5237} &
  0.4951 &
  0.5027 &
  0.4151 &
  0.3772 &
  \textbf{0.5965} &
  0.5677 &
  0.5464 &
  0.4302 &
  0.4277 \\ \cmidrule(l){1-2} \cmidrule(l){3-7} \cmidrule(l){8-12}
\multirow{5}{*}{\textbf{csCG-TTB}} &
  \textbf{POS} &
  \textbf{0.5740} &
  0.5416 &
  0.5531 &
  0.5024 &
  0.4526 &
  \textbf{0.6012 (6)} &
  0.5947 (6) &
  0.5759 (5) &
  0.5516 (8) &
  0.4817 (4) \\
 &
  \textbf{STDP} &
  0.1029 &
  0.1184 &
  \textbf{0.1476} &
  0.0716 &
  0.0716 &
  0.2431 (11) &
  0.1085 (1) &
  \textbf{0.2556} (4) &
  0.0716 (12) &
  0.1467 (7) \\
 &
  \textbf{GCM} &
  0.5931 &
  0.5832 &
  0.5627 &
  0.5605 &
  0.5717 &
  0.6300 (11) &
  0.6065 (8) &
  0.5811 (2) &
  0.5852 (4) &
  0.5875 (7) \\
  \cmidrule(l){2-2} \cmidrule(l){3-7} \cmidrule(l){8-12}
 &
  \textbf{Average} &
  \textbf{0.4233} &
  0.4144 &
  0.4211 &
  0.3782 &
  0.3653 &
  \textbf{0.4914} &
  0.4366 &
  0.4709 &
  0.4028 &
  0.4053 \\ \bottomrule 
\end{tabular}
}
\caption{The weighted-F1 scores for the last layer and the best layer (layer-number mentioned in parenthesis) for the layer-wise probing experiments with the Tamil language data.}
\label{tab:results-last-best-tamil}
\end{table*}
\FloatBarrier

\section{Experiments with Tamil Language}
\label{sec:tamil-appendix}

Due to limitations from the original TTB Tamil Universal Dependencies treebank, we do not  cover certain grammatical-gender-based methods and experiments with Tamil language in the current work.\footnote{This includes the controlled generation of colorless green sentences with gender-balancing, and the Subject-Verb Agreement (SVA) task.} Here, we report all the other results for the CG-TTB and csCG-TTB treebank datasets.

\subsection{Probing Results}
\label{sec:appendix-tamil-results}
While comparing the performance between the models, all the observations and inferences discussed in Section~\ref{sec:probing} also hold true for the Tamil language data as seen in Table~\ref{tab:results-last-best-tamil}. Overall, we observe a relatively lower performance in the CG-TTB and csCG-TTB datasets (Table~\ref{tab:results-last-best-tamil})  as compared to CG-HDTB and csCG-HDTB datasets respectively (Table~\ref{tab:results-last-best}).

\subsection{Localization of Syntactic Knowledge}
\begin{figure}[htbp]
  \includegraphics[width=0.9\columnwidth, left]{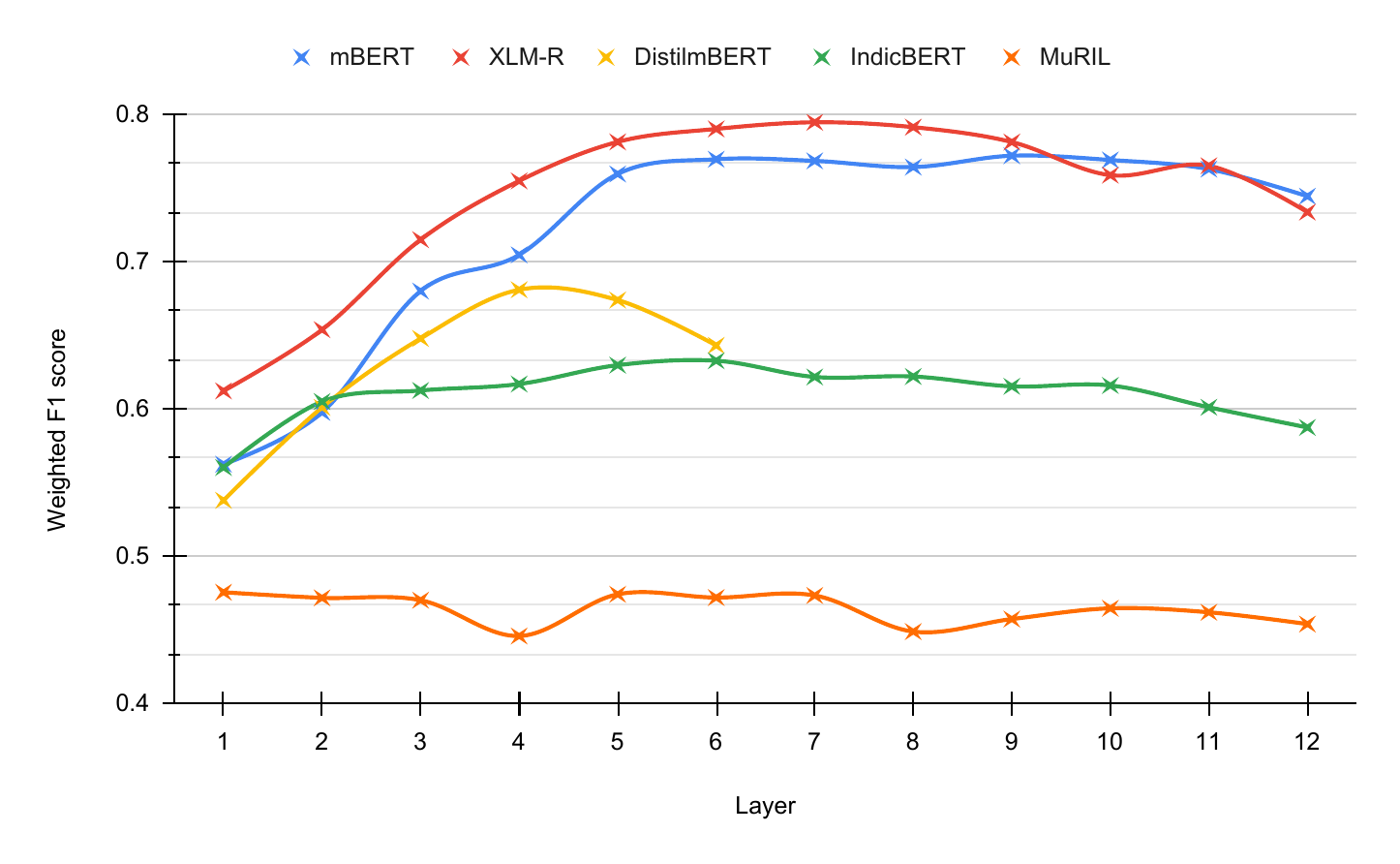}
  \caption{Weighted F1 scores for the layer-wise probing experiments with the CG-TTB test set for the POS task.}
  \label{fig-cgi-tamil-postag}
\end{figure}

\begin{figure}[htbp]
  \includegraphics[width=0.9\columnwidth, left]{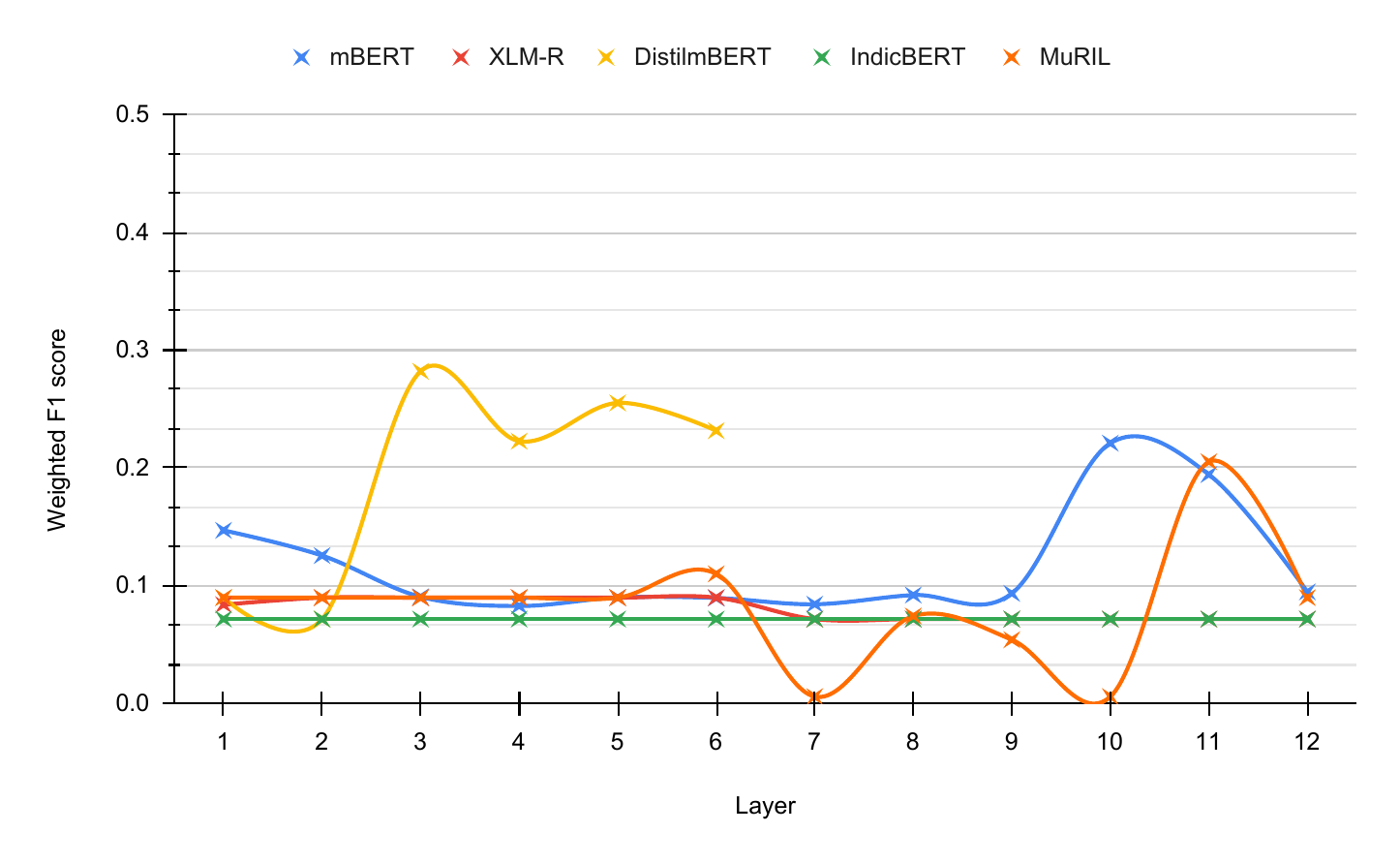}
  \caption{Weighted F1 scores for the layer-wise probing experiments with the CG-TTB test set for the STDP task.}
  \label{fig-cgi-tamil-treedepth}
\end{figure}

\begin{figure}[htbp]
  \includegraphics[width=0.9\columnwidth, left]{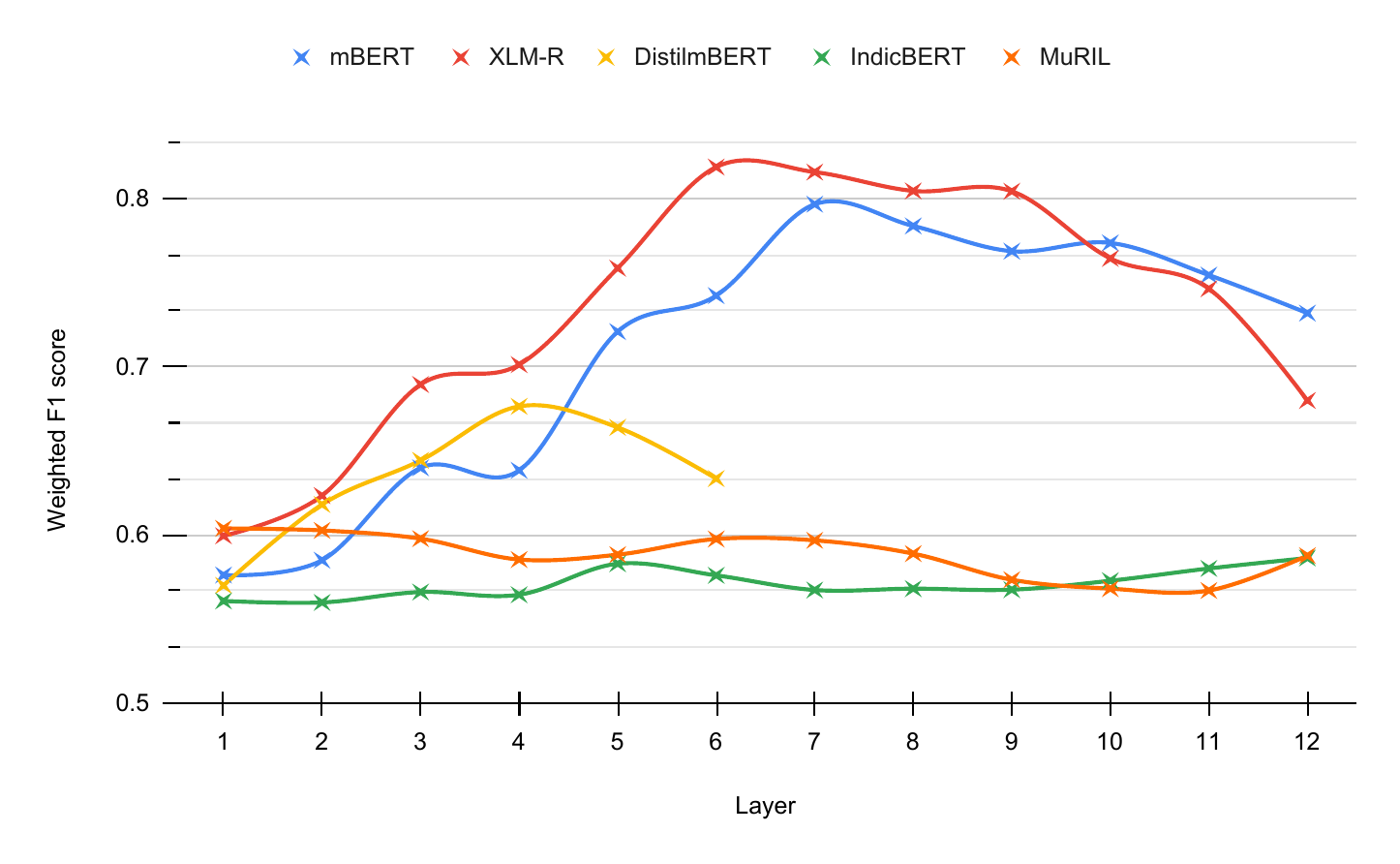}
  \caption{Weighted F1 scores for the layer-wise probing experiments with the CG-TTB test set for the GCM task.}
  \label{fig-cgi-tamil-case}
\end{figure}

All the observations and inferences discussed in Section~\ref{sec:localization} also hold true for the Tamil language data as seen in Figure~\ref{fig-cgi-tamil-postag}, Figure~\ref{fig-cgi-tamil-treedepth}, and Figure~\ref{fig-cgi-tamil-case}. Due to a relatively lower performance on Tamil language data, all the figures are shifted downwards with almost similar trends and structures.

\end{document}